# Automatic speech recognition for the Nepali language using CNN, bidirectional LSTM and ResNet

Manish Dhakal, Arman Chhetri, Aman Kumar Gupta, Prabin Lamichhane, Suraj Pandey, Subarna Shakya
*Department of Electronics and Computer Engineering*
*Pulchowk Campus - Institute of Engineering, Tribhuvan University*
Lalitpur, Nepal
{074bct521.manish, 074bct506.arman, 074bct503.aman, 074bct523.prabin, 074bct547.suraj}@pcampus.edu.np,
drss@ioe.edu.np

*Abstract*—This paper presents an end-to-end deep learning model for Automatic Speech Recognition (ASR) that transcribes Nepali speech to text. The model was trained and tested on the OpenSLR (audio, text) dataset. The majority of the audio dataset have silent gaps at both ends which are clipped during dataset preprocessing for a more uniform mapping of audio frames and their corresponding texts. Mel Frequency Cepstral Coefficients (MFCCs) are used as audio features to feed into the model. The model having Bidirectional LSTM paired with ResNet and one-dimensional CNN produces the best results for this dataset out of all the models (neural networks with variations of LSTM, GRU, CNN, and ResNet) that have been trained so far. This novel model uses Connectionist Temporal Classification (CTC) function for loss calculation during training and CTC beam search decoding for predicting characters as the most likely sequence of Nepali text. On the test dataset, the character error rate (CER) of 17.06 percent has been achieved. The source code is available at: https://github.com/manishdhakal/ASR-Nepali-using-CNN-BiLSTM-ResNet.

*Index Terms*—Nepali Speech Recognition, Residual Network, Convolutional Neural Network, Bidirectional Long Short Term Memory

## I. Introduction

Automatic Speech Recognition (ASR) is a technology in the field of computer science that transcribes the speech to its equivalent written text. ASR has captured the attention of the artificial intelligence community over the last few decades. ASR has numerous applications in the healthcare system, banking, marketing, home automation, and many more.

Classic speech recognition models based on Gaussian Mixture Models (GMM) [1] and Hidden Markov Models (HMM) [2] used to be the gold standard for ASR models. End-to-end (E2E) ASR has reached new heights after the advancements in the field of Deep Neural Networks (DNNs), Recurrent Neural Networks (RNNs), and extended computing power. Recent trends show that ASR models are mostly based on variants of RNN: conventional RNN, Long Short Term Memory (LSTM), or Gated Recurrent Unit (GRU) [3], [4], [5]. E2E technique has resulted in the development of new techniques such as Connectionist Temporal Classification (CTC) [6], Sequence Transduction [7], and Attention-Based Sequence-To-Sequence (seq2seq) [8] learning. Encoder-Decoder [9] (a sequence-to-sequence modeling technique) engages an RNN acoustic encoder to encode the sequential audio into a fixed-length vector, which is then decoded by another RNN into a text sequence. Convolutional Neural Network (CNN) has been used in several speech recognition models to enhance predictive ability [10]. Some systems combine CNN with RNNs to boost the efficiency of speech processing [11]. In some of the ASR models, the networks have been designed with the utilization of the Residual Network (ResNet) technique [12], [13]. Pretraining or representation learning with self-supervised speech recognition models has been on the rise [14], [15]. Lately, the attention mechanism [16], [17] has been coupled with the CTC model to produce a satisfactory result [18]. Mel Frequency Cepstral Coefficient (MFCC) [19] has been around for quite some time. In the ASR field, MFCC is a powerful feature extraction technique that uses power spectra. It generates a feature vector from multiple audio frames.

The study of Nepali language ASR is still inadequate. A few pieces of research have been discovered that used neural networks (NNs) for ASR of the Nepali language [20], [21], [22]. These papers employed character-level prediction with CTC. For sequence mapping of audio to text, [20] used Bidirectional LSTM whereas, [21] and [22] employed GRU. Same as in our models, [22] used the OpenSLR dataset [23] for training and testing.

In this paper, with effective data cleaning, a powerful data preprocessing method, and an optimal neural network, Nepali texts have been transcribed by using techniques mentioned in Section II. For data cleaning, the silent gaps have been clipped from both ends of the audios using the algorithm defined in Section II-A. MFCC, as a robust feature extraction technique, has been employed in our model as shown in Section II-A. The performance of variants of neural networks (NNs) which have different combinations of one-dimensional convolution neural network (1D-CNN) [24], ResNet [25], and RNN have been evaluated to obtain the optimal NN. The CTC function [6] is used to calculate the model's loss as it does not require the actual alignment between the input audio and the output characters sequence as explained in Section II-B4. The final output of the NN is the probability distribution of unique Nepali characters. To generate the sequence of characters from the probability distribution of the model, a modified beam search called CTC beam search [26] is used. Fig. 1 shows the training flow whereas the Fig. 2 shows the inference pipeline of the trained model. After observing from the Section IV, the variant having 1D-CNN, ResNet, and bidirectional LSTM (BiLSTM) proves to be the best neural network model among the many models that have been trained in this paper. ResNets increase the model's complexity and expressiveness. The model's prediction is attained at the character level. In Table II, the promising transcriptions obtained from that model can be seen.

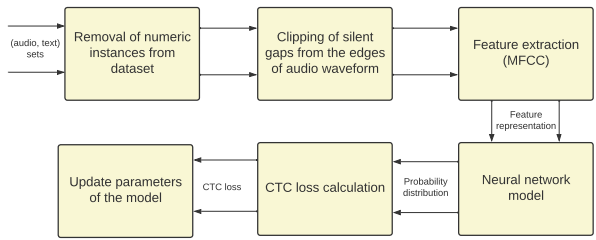

Figure 1. Training flow of the ASR model.

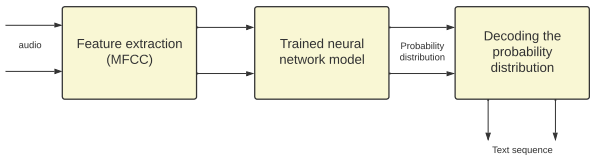

Figure 2. Inference flow of the ASR model.

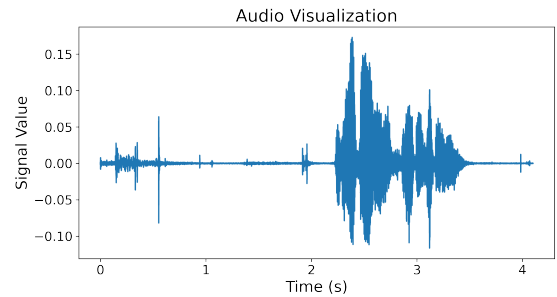

Figure 3. Waveform of audio XYZ with silent gaps at the both ends.

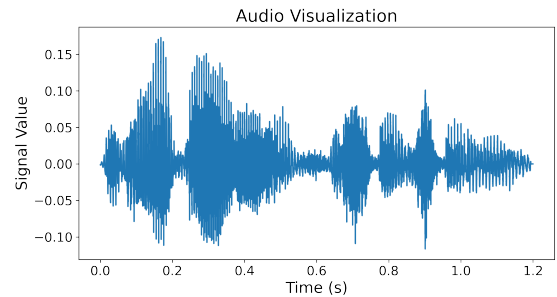

Figure 4. Waveform of audio XYZ after clipping of silent gaps.

## II. Methodology

### A. Data Acquisition and Preprocessing

The dataset [23] used in our model is a collection of 157,905 audio clips with 527 unique speakers, sampled at 16KHz. The source of the dataset is OpenSLR, which hosts speech and language resources, such as training corpora and software related to speech recognition. Initial cleansing performed on the dataset was the elimination of the numeric transcriptions (such as १९२४, २३०, ३, etc.). Since this type of data had minimal instances present, their inclusion while training would degrade the overall performance of the model. After the removal of those instances, 143.6 hours of 148,188 audio clips are left, which is the foundation of training and testing performed on our designed model.

The majority of the audio clips in the dataset have large silent gaps at both ends. For the clipping of silent gaps, sliding window processing has been implemented on the absolute intensities of the waveform. The processing starts from both ends and moves toward the center. In that processing, the local mean of the window segment of the audio wave is calculated and compared with the audio's global mean. If the local mean is smaller than the global mean, the window segment is considered to contain insignificant data, thus the segment is clipped from the original audio. Algorithm 1 clarifies the above-mentioned steps. Fig. 3 shows the original waveform of an audio signal which is clipped to form the shorter audio of Fig. 4. This prepossessing step reduces the dataset to 44.3% (63.6 hours) of the original size, resulting in faster individual epoch training. After this clipping process, the alignment between the input audio and the output characters is still unknown but more uniform. Uniform alignment eventually helps us to achieve faster saturation of CTC loss to the minimal value. The CTC loss has been explained in Section II-B4.

During preprocessing, the audio data were processed using the powerful feature extraction mechanism, which generates coefficients known as Mel Frequency Cepstral Coefficients (MFCCs) [19]. The feature extraction mechanism includes six stages. Pre-emphasis enhances higher-frequency components, framing divides samples into segments, and windowing multiplies samples using a scaling function to smooth the signal near edges. The resultant signal is subjected to a Short Time Fourier Transform, which is then applied to the mel filter banks. Mel filter banks have the collection of bandpass filters over the mel scale and each band carries a corresponding signal decomposed from the original audio signal. Mel scale is the measure of frequency based on the nonlinear perception of the pitch by human ears. The MFCCs are finally obtained by transformation of mel scale features to the time domain using the Discrete Cosine Transform. The matrix obtained as

**ALGORITHM 1**: Clipping of silent gaps from both ends

$wav \leftarrow sampled\ audio\ signal$
$\Delta \leftarrow appropriate\ window\ length$
/* In our code, $\Delta = 500$ for 16KHz sampling rate*/

**INPUT:** $wav, \Delta$
**PROCESS:**
$wavAvg \leftarrow Average(|wav|)$
$N \leftarrow Length(wav)$

/* Removing the silent gap from the start */
**for** $idx = 0,\ \Delta,\ 2\Delta,\ \ldots, N-\Delta$ **do**
   $win \leftarrow wav[idx : idx + \Delta]$
   $winAvg \leftarrow Avergae(|win|)$
   **if** $winAvg > wavAvg$ **then**
     $wav \leftarrow wav[idx :]$
     **break**
   **end if**
**end for**

/* Removing the silent gap from the end */
**for** $idx = N - \Delta,\ N - 2\Delta, \ldots, 0$ **do**
   $win \leftarrow wav[idx : idx + \Delta]$
   $winAvg \leftarrow Avergae(|win|)$
   **if** $winAvg > wavAvg$ **then**
     $wav \leftarrow wav[: idx]$
     **break**
   **end if**
**end for**

**OUTPUT:** $processed\_wav \leftarrow wav$

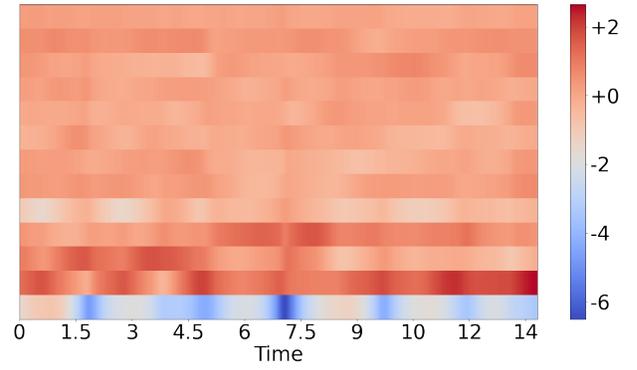

Figure 5. Normalized MFCCs of the audio XYZ.

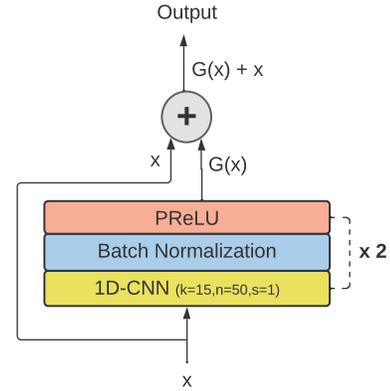

Figure 6. Residual learning block (ResNet) consisting of Batch Normalization, Parametric ReLU activation layer, and of 1D-CNN with kernel size (k), number of feature maps (n), and stride (s) used in the CNN layer.

MFCCs gives us features of the sound based on power spectra. MFCCs are quite favorable with RNN, DNN, and CNN for signal processing [27]. Fig. 5 shows the normalized MFCCs of an audio.

For human speech, 13 mel scales are sufficient for extracting features from the signal [28], as implemented in our data preprocessing. The equation involved in the calculation of the mel scale from the frequency in Hertz (f) is given by:

$$Mel(f) = 2595 * log(1 + \frac{f}{700}) \quad (1)$$

*B. Machine Learning Components*

*1) ResNet:* As the depth of any neural network model increases, so does the model's susceptibility to degradation. One may argue that the degradation is the result of overfitting, but that is not the case. As the network depth grows, the degradation is driven by a larger training error. Thereafter, the accuracy of the model gets saturated and starts to degrade rapidly. Residual Network (ResNet) was introduced to address this issue [25]. For such deep networks, it is easier to optimize the residual learning framework than the conventional neural network. Originally, ResNet was developed for endorsing computer vision, but it is adequately applicable in speech processing.

ResNet can be realized using shortcut connections. Shortcut connections take a tensor and add it to the output of the NN layers that are stacked on top of it. The residual block for our model has an output of $G(x) + x$, where $x$ is the block's input and $G(x)$ is the output of stacked layers. The stacked layers must be designed so that $x$ and $G(x)$ have the same shape. Fig. 6 helps us to visualize the above-mentioned steps.

For the implementation of residual block in our model, there exists 1D-CNN [24] and Batch Normalization (BN) [29] as in Fig. 6. 1D-CNN extracts localized features. For model training, BN adds stability and speed to gradient descent. It speeds up DNN training by reducing internal covariate shifts [29]. It normalizes the CNN layer's output vector by using the mean and the variance from the current batch. The normalized values are scaled and shifted by two learnable parameters: gamma and beta respectively. The activation function for the output of the BN layer is Parametric ReLU (PReLU) [30]. Furthermore, the output of the PReLU function is added with the input to the residual block. Finally, a residual learning block is fully implemented in our model.

*2) 1D-CNN:* Convolutional Neural Networks (CNN) [31] provide a better-localized feature extraction mechanism with fewer trainable parameters than pure Multi-Layered Perceptrons (MLP). CNN is based on the convolution operation, which in the field of Digital Signal Processing (DSP), is a standard mathematical operation on two signals. CNN is the result of combining convolution operation with a neural network. In 1D-CNN, convolution takes place in the temporal direction of the signal and uses trained weighted filters (kernels) to extract localized features [24]. The obtained feature mapping would be fed into the neural network.

*3) RNN:* RNN is a neural network that uses the previous time step's output as input for the current time step. It proves to be useful when working with sequential data such as speech and text. Two RNN variants are proposed for use in our ASR model: LSTM and GRU.

*a) LSTM:* The LSTM [32] is a gradient-based neural network that solves the problem of vanishing or exploding gradients existing in the classic RNN. Large interval contexts are easily learned in LSTM by enforcing constant error flow within special units via constant error carousels [32]. Gates control the cell state, which is transferred as information from one cell to the next. The gates involved in calculations are the input gate, the output gate, and the forget gate. As a result, LSTMs can propagate useful contextual information while discarding irrelevant information. LSTMs are more effective than conventional RNNs and GMMs for acoustic modeling in speech recognition [33], [34].

*b) GRU:* GRU [35], which has fewer parameters than LSTM, is another gated mechanism for contextual information flow in RNN. They do not have a distinct cell state, only a hidden state. GRU training has a lower time and space complexity than LSTM. In general, the significance of GRU and LSTM is comparable. GRUs have two gates: the reset gate and the update gate. GRUs are more efficient than LSTMs in some cases for speech recognition and speech signal modeling [5], [35].

*4) CTC:* Every supervised learning algorithm is characterized by the task of functionally mapping input to the output. In a speech-to-text conversion, the sequence of speech data is mapped to a sequence of transcripted text. People have varying rates of speaking, so the alignment between the input audio and the output text is unknown in the training dataset. Also, it is practically impossible to manually align each output character to its exact location in the audio while dealing with thousands of audio signals. CTC loss [6] helps us to deal with not knowing the alignment between the input audio and the output characters. Alignment-free loss value is computed by introducing a token called blank token during training and inferring.

The task is to train a model to calculate a conditional probability $p(Y|X)$ as in (2) where $X$ and $Y$ are input and output sequences respectively. A probability distribution $p_t(a_t|X)$ over the vocabulary is obtained by feeding the audio features into a deep learning model such as RNN. The objective function is the sum of probabilities of all possible valid sequences.

Mathematically, [26]

$$p(Y|X) = \sum_{A \epsilon A_{x,y}} * \prod_{t=1}^{T} p_t(a_t|X) \qquad (2)$$

where $A_{x,y}$ is the valid alignment of $Y$ given $X$

*C. Components Infusion*

For our candidate models, two prominent approaches have been used: a sequence model with ResNet and a sequence model without ResNet. Multiple residual learning blocks of Fig. 6 are stacked together in the ResNet model. The last residual block's output is used as input for the block of stacked multiple RNN layers, as shown in Fig. 7. When multiple RNN layers are stacked together, the model can provide multiple levels of feature abstraction to generalize the existing pattern in the data. In the plausible models, either the LSTMs or GRUs have been used as the RNNs variations. Bidirectional RNN (BiRNN) [36], [37], an RNN extension, is used in some of the models that have been trained. In both BiRNNs: BiLSTM and BiGRU, the character prediction at a time frame is dependent on both previous and next information from the input sequence. To avoid overfitting while training the model, dropout techniques [38] for RNNs are introduced. Following that, the RNN layers are linked to the neural network's dense layers. Finally, the output is drawn as the model's softmax output. The softmax output contains the probability distribution of 66 unique characters. The CTC loss is obtained as an error margin between the targeted transcriptions and softmax outputs. During training, the loss is backpropagated within the network to update the model's parameters. The model with the lowest CTC loss is chosen as our final model after the numerous updates in the parameters.

During prediction, subsequent softmax outputs of the model must be decoded to obtain a sequence of characters. The greedy search algorithm decodes a character with the highest probability from the softmax vector. This method is quick, but the output text may be inaccurate. The CTC beam search decoding method tries to find out the most probable output by considering the top-n alignments in each time step [26]. It achieves this in reasonable time complexity using dynamic programming and limiting the beam width, thus CTC beam search was used to decode the softmax output.

III. EXPERIMENTAL SETUP

The proposed model was trained using the OpenSLR[1] dataset [23] by removing all of the audio-sentence pairs that contain any Nepali numerals transcription. The dataset's silent gaps from ends were clipped by using the Algorithm 1. The dataset was divided into two sets: training (95%) and testing (5%) datasets. The audio data was sampled at 16KHz considering each frame having a size of 160 (ie. each second of audio had 100 frames). The features were extracted from the audio file as MFCC with 13 mel scales and hop size of 40. This resulted in the decreased frame vector of dimension 52 as shown in Fig. 7 and a single input frame was processed to predict a token. Each of the frames gave the softmax output for the character among the 66 unique tokens (63 Nepali characters "not including the numeric digits", 1 padding token, 1 unknown token, and 1 blank token). The loss was computed as the CTC loss (as in Fig .8) which gives the gradient to update the weights of the model. The optimizer for the gradient descent was used as Adam [39] where the learning rate was 0.001, $\beta_1$ is 0.9, $\beta_2$ is 0.999, and decay is 0.

---
[1]https://openslr.org/54/

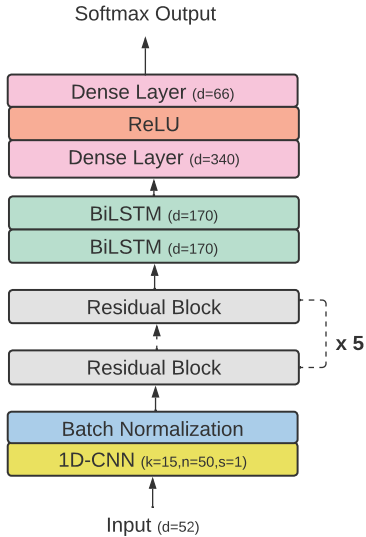

Figure 7. Architecture of proposed ASR model consisting of 5 residual blocks (as in Fig. 6) and 2 BiLSTM layers with the given hyperparameters as ( kernel size (k), number of output feature map (n), and stride (s) for CNN layer ), and dimension of other layers (d).

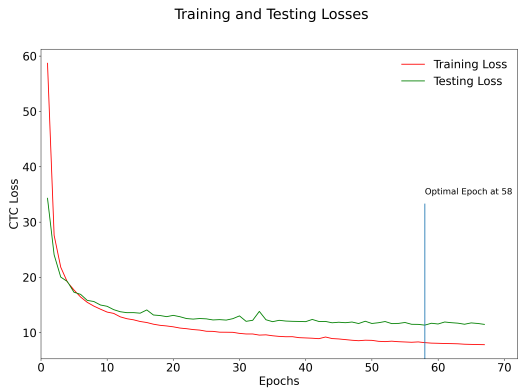

Figure 8. Training and Testing loss of the proposed sequence Model (CNN + ResNet + BiLSTM) from Fig. 7.

Table I
MODELS AND THEIR CER ON TEST DATA (5% OF TOTAL DATA)

| Model | Test Data CER ↓ | # Params |
|---|---|---|
| **Our trained models** | | |
| BiLSTM | 19.71% | 1.17M |
| 1D-CNN + BiLSTM | 24.6% | 1.55M |
| 1D-CNN + ResNet + BiGRU | 29.6% | 1.30M |
| **1D-CNN + ResNet + BiLSTM** | **17.06%** | **1.55M** |
| 1D-CNN + ResNet + LSTM | 30.27% | 0.88M |
| **Other** | | |
| 1D-CNN + GRU [22] | 23.72% | - |

Variants of sequence models which have been mentioned in Table I were used to train and test those MFCC features for determining the best model among them. The proposed model has 1.55 million parameters and a dropout in the BiLSTM layers of 25%. A single forward-backward pass over the entire training dataset (an epoch) took the time of roughly 20 minutes with the batch size of 80. The training was performed up to 58 epochs in the GPU of the NVIDIA Tesla T4 system.

During the inference of trained models, MFCC features were extracted from audio, which were fed into the model to give the softmax output. The softmax output was then decoded by using CTC beam search decoding.

IV. RESULT AND ANALYSIS

The CTC loss as shown in Fig. 8 is calculated on the training and test datasets to determine the model's efficiency for each training epoch. A lower value of the CTC loss indicates that the decoded sequence of characters is nearer to its targeted transcription.

Character error rate (CER) has been used as the evaluation metric for being more comprehensible. After beam search decoding of the softmax output, those blank tokens and redundant characters are removed to get the understandable prediction of the characters' sequence. The rate of incorrect prediction of characters by the model is computed as character error rate (CER). CER on the test dataset has been used as an evaluation metric for the distinct models that have been trained so far, as illustrated in Table I.

After observing Table I, it has been discovered that the model combining 1D-CNN and BiLSTM has a higher error than the standalone BiLSTM model. The phenomenon is caused as the depth of the neural network grows and the model becomes more prone to degradation [25]. The training error gets saturated at a higher value and the model starts to degrade rapidly. To address this issue, ResNet has been embedded into that model (1D-CNN + BiLSTM). Thus, the best model that has been designed so far is a combination of ResNet, 1D-CNN, and BiLSTM. On the unseen test dataset, a CER of 17.06% (82.94% of character accuracy rate) has been achieved with that model. So far, the mentioned ASR model outperforms all other sequential models in terms of test accuracy for this OpenSLR dataset.

Table II illustrates some examples of each of our trained models' prediction on a dataset[2] gathered verbally from our associates and friends.

---
[2]The crowdsourced data is outside of the training and testing datasets.

Table II
MODELS AND THEIR TRANSCRIPTION OF CROWDSOURCED SPEECHES

| Actual Transcription | Model | Predicted Transcription |
|---|---|---|
| मलाई गीत गाउन मनपर्छ | BiLSTM | मलाई गीत गाउन भन्पर्छ |
| | 1D-CNN + BiLSTM | मलाई जितगाउन मनुपर्छ |
| | 1D-CNN + ResNet + BiGRU | माल दिनगयाउनु हुन पछ |
| | **1D-CNN + ResNet + BiLSTM** | मनाई जीत गाउन मनपछ |
| | 1D-CNN + ResNet + LSTM | मालाई जित ल्‌उनुहुनपर्छ |
| सुन्ने समयमा बोल्नु चाहनु हुने माननीय सदस्यहरूको लागि पाँच मिनेटको समय निर्धारण गरिएको छ | BiLSTM | सुन्य समयमा बूल्न चाहने हुनेमा मानीय सदश्यहरूका लागि आफुकिो सबय निर्धारयण गरिएको छ। |
| | 1D-CNN + BiLSTM | सुन्ने समयमा बूरम चहाहदे उनेमा मानगीय शरस्यहरूका लागि आँचुरेको समैनिर्धारान गरिएको छ |
| | 1D-CNN + ResNet + BiGRU | सुन्ने समयमाँ बुल्रजहाने हुनेमा मानगी सदष्हरूका लागि पासु-नेको समई नेवार गरिएको छ |
| | **1D-CNN + ResNet + BiLSTM** | सुन्ने समयमा बोल्न चाह ने हुने मा मानवी सदस्यहरूका लागि भाँसुनेको सम निर्धारण गरि-एको छ। |
| | 1D-CNN + ResNet + LSTM | सुन्ने समयमा पनम चाहँने उने मामारणी सदस्यहरूका लागि भासीरेको समानयरारण गरि-एको छ |
| नेपालको राजधानी काठमाडौँ हो | BiLSTM | पालको राजधानी काठमाडौँ हो |
| | 1D-CNN + BiLSTM | नेपालको राजधारी काठमाडौँ |
| | 1D-CNN + ResNet + BiGRU | नेपालको रास्ताडीकाठ माटो हो |
| | **1D-CNN + ResNet + BiLSTM** | नेपालको राजधानी काठमाडौँ हो |
| | 1D-CNN + ResNet + LSTM | नेपालको राजधानीकाठमाडौँहो |
| तिमीलाई ठुलो भए पछि के बन्ने मन छ | BiLSTM | तिमीलाई खुलभएपरछि कय भन्नी भन्छ |
| | 1D-CNN + BiLSTM | तिवीलाईखुनभएपछि कौ भन्ने मन्छ |
| | 1D-CNN + ResNet + BiGRU | तिमीलाईखलोभयपसित् भन्ने भन्छ |
| | **1D-CNN + ResNet + BiLSTM** | तिमीलाई ठुनभएपछि केवनी मन छ |
| | 1D-CNN + ResNet + LSTM | तिमीलाई ठुलभए पछि के मन्ने मन्छ |

## V. CONCLUSION

We have trained several models and discovered that ResNet combined with 1D-CNN and BiLSTM produces the best result. ResNet assisted us in overcoming the limitation of early saturation of the CTC loss value. Because of the efficient data cleaning process, the alignment between the audio frames and their corresponding characters is improved.

ACKNOWLEDGMENT

We express our gratitude towards the Department of Electronics and Computer Engineering, IOE Pulchowk Campus for helping us conduct the research as an academic project. We convey our profound appreciation to Dr. Basanta Joshi for providing us with his support and the crowdsourced data for inspection of our model.